\documentclass[11pt]{article}

\usepackage[a4paper,margin=1in]{geometry}
\usepackage{amsmath}
\usepackage[T1]{fontenc}
\usepackage{lmodern}
\usepackage{microtype}
\usepackage{booktabs}
\usepackage{multirow}
\usepackage{graphicx}
\usepackage{threeparttable}
\usepackage{tikz}
\usetikzlibrary{positioning,arrows.meta,fit,calc}
\usepackage[authoryear,round]{natbib}
\usepackage{setspace}
\usepackage[hidelinks]{hyperref}

\graphicspath{{./}}

\title{\textbf{Whose hotel does the AI recommend? An algorithm audit of
reputation signals in LLM-assisted hotel selection}}

\author{
  Mirza Samad Ahmed Baig\thanks{Equal contribution. Corresponding author:
  \texttt{Mirzasamadcontact@gmail.com}.}\;\thanks{Fandaqah, Al Khobar, Saudi
  Arabia.}\textsuperscript{,}\,\footnotemark[3] \and
  Syeda Anshrah Gillani\footnotemark[1]\;\thanks{Hamdard University, Karachi,
  Pakistan.} \and
  Asher Ali\footnotemark[2]
}
\date{\today}

\begin{document}

\maketitle

\begin{abstract}
\noindent Travelers increasingly ask large language model (LLM) assistants which
hotel to book, making these systems gatekeepers of property visibility---yet what
moves their recommendations is undocumented. We conduct a pre-specified algorithm
audit using a randomized choice-based conjoint: across personas, prompt templates,
and twelve open-weight and proprietary models, assistants choose among five hotels
whose guest rating, review volume and recency, management response, chain
affiliation, price, eco-certification, and list position are independently
randomized. We estimate the average marginal component effect of each signal on
the probability of recommendation. Guest rating and price dominate (a top rating
raises selection by 31.6 percentage points; a high price lowers it by 30.0),
reproducing human valence-and-price primacy but over-weighting eco-certification
and ignoring management response. List position---a content-free artifact---shifts
recommendations causally, worth about \$12 per night. Stated reasons track
revealed weights imperfectly. The findings ground generative engine optimization
and the accountability of AI infomediaries in causal evidence.
\end{abstract}


\bigskip

\section{Introduction}\label{sec:intro}

The most important shop window a hotel has is no longer a page of search
results---it is increasingly the context window of a conversational assistant.
When a traveler types ``which of these hotels should I book?'' into a
general-purpose large language model (LLM), the assistant does not return a
ranked list to be browsed and filtered; it returns a recommendation, often a
single property, accompanied by a short justification. Industry tracking
indicates that roughly forty percent of United States travelers used
generative-AI tools to plan a trip during 2025, an increase of about eleven
percentage points year over year, and that the share of travelers who begin
trip planning at a conventional search engine has fallen sharply while the share
beginning at a generative-AI platform has more than doubled \citep{phocuswright2025}.
Forward-looking surveys point the same direction: a majority of travelers now
expect to use AI for trip planning within the year, and among prospective users,
obtaining recommendations is the single most common intended use. Since major
assistants began enabling hotel search and booking directly inside the chat in
2025, the LLM has moved from advising on the periphery of the funnel to
occupying its decisive center---the point at which one property, and not its
competitors, is surfaced to a ready-to-book traveler.

This relocation of the decision turns the assistant into an \emph{algorithmic
infomediary}: a gatekeeper whose selection function determines which eligible
properties a traveler ever sees \citep{metaxa2021}. For an individual hotel, the
stakes are concrete and asymmetric. A property that the assistant systematically
declines to surface is, for that growing population of travelers, effectively
invisible---and, unlike a low organic search ranking, the hotel cannot observe
the slot it failed to win, cannot see its competitors' treatment, and is given
no account of why. Suppliers have nonetheless begun to act. A nascent industry
practice, \emph{generative engine optimization}---the deliberate shaping of
content to win inclusion and prominence in generative-engine answers, advanced
as the successor to search engine optimization---already treats the LLM's
selection behavior as a surface to be engineered \citep{aggarwal2024geo}. Yet
this practice is being sold and adopted with essentially no causal evidence base:
the inputs that actually move an assistant's recommendation are undocumented, and
the field has no estimate of \emph{which} reputation signals shift the machine's
choice, in which direction, or by how much. Optimization advice is therefore
running well ahead of measurement.

We argue that the question with both theoretical and managerial bite is a
\emph{supply-side} one: among the reputation signals a hotel can actually
manage---its average guest rating, the volume and recency of its reviews,
whether management is seen to respond, its brand or chain affiliation, its price,
and its third-party eco-certification---which causally change the probability
that an LLM recommends the property, and with what relative weight? This question
inherits decades of human evidence as a benchmark. The electronic word-of-mouth
(eWOM) literature has established not only that these cues move human booking
behavior but, crucially, their \emph{relative ordering}: review valence (rating)
is consistently the dominant lever, far outweighing review volume, with
management response, brand, price, and green certification playing measured
secondary roles \citep{yang2018,proserpio2017,assaker2023}. Whether the
gatekeeper that increasingly stands between hotels and travelers reproduces that
hard-won ordering, or reweights the cues in idiosyncratic and commercially
consequential ways, is unknown.

We answer this question with a randomized, choice-based conjoint
\emph{algorithm audit} \citep{sandvig2014,metaxa2021,hainmueller2014}. In each
choice set, an LLM travel assistant is asked to recommend one of five synthetic
hotel cards whose reputation attributes---rating, review volume, review recency,
management response, chain affiliation, price, and Green Key eco-certification,
together with the list position of each card---are independently randomized.
Because the attributes are randomized, the average marginal component effect
(AMCE) of each signal on the probability of recommendation is causally
identified, in interpretable percentage-point units, and is comparable across
models and decoding temperatures \citep{hainmueller2014,swait1993}. We run the
audit over three traveler personas and nine prompt paraphrases, collecting
3{,}024 main-arm choice sets per model under common stimuli, across a panel of
twelve models---four open-weight systems run locally (Llama-3.2-3B, Qwen-2.5-3B,
Phi-3-mini, and Mixtral-8$\times$7B) and eight proprietary models accessed by API
(OpenAI GPT-4o-mini; Google Gemini 1.5 Pro, 2.0 Flash, and 2.0 Pro; and four
Anthropic Claude models)---for a total of more than sixty thousand model calls
across all arms. We deliberately scope the
study to the \emph{selection stage}---the assistant's choice among a fixed,
already-retrieved set of candidates---rather than the retrieval or multi-turn
negotiation stages; this isolates the gatekeeping decision under clean
experimental control and is, we argue, the stage at which a hotel's managed
reputation signals are most directly in contention. The full design, hypotheses
(H1--H12), and analysis plan were specified and cryptographically hashed before
any confirmatory data were collected, and the complete design, prompts, code, and
raw model outputs are archived for inspection.

Our findings can be summarized as follows. The pooled panel is moved most by
guest rating (a top rating raises recommendation probability by 31.6~pp) and by
price (a high price lowers it by 30.0~pp), with eco-certification (+11.6~pp) and
review volume (+8.3~pp) as secondary positive signals; chain affiliation carries a
small penalty ($-1.8$~pp), and management response---a cue the optimization
industry actively promotes---has no detectable effect (+0.1~pp, statistically
equivalent to zero). The revealed importance ordering reproduces the human
primacy of rating and price and the valence-over-volume pattern of the eWOM
benchmark, but departs from it by elevating eco-certification to the third-ranked
signal while all but ignoring management response and brand. List position
causally affects recommendation: pooled, the first slot carries a modest advantage
worth about \$12 per night (roughly a tenth of a full rating step), but this masks
extreme heterogeneity, with one model showing a first-position advantage an order
of magnitude larger. Across personas, attribute weights shift in the
theoretically expected directions---price sensitivity is highest for the budget
family, eco-certification most valuable to the eco-conscious couple, and the chain
penalty sharpest for that same persona while vanishing for the business traveler.
The models' stated reasons correspond positively but imperfectly with their
revealed weights (Spearman $+0.59$ to $+0.85$): they act on list position and
review volume without naming them, and over-cite brand relative to its near-zero
influence. These patterns are robust across the nine prompt paraphrases, between
decoding temperatures 0.7 and 0.0, across card and grounded-snippet layouts, and
across choice-first and reason-first field orders.

The paper makes three contributions. First, it provides the first causal
estimates of how managerially actionable reputation signals weigh in an LLM's
hotel recommendation, and it positions those estimates against published human
benchmarks---thereby characterizing not just \emph{whether} the algorithmic
gatekeeper attends to reputation, but how its reputation function \emph{diverges}
from the human eWOM evidence against which hotels have long been managed. Second,
it prices the assistant's own positional distortion in units a manager or
platform can act on: by randomizing list position independently of content, we
convert the position effect into a price-equivalent (\$/night) and a
rating-step equivalent, moving beyond the existing computer-science finding that
LLM recommenders are order-sensitive \citep{bito2025,hou2024,ren2024} to a
quantified, managerially legible magnitude. Third, by comparing what each model
\emph{says} drives its choice with what \emph{actually} does, it offers evidence
on the stated-versus-revealed transparency of these systems as accountable
intermediaries, and it converts that evidence into the foundation for an
\emph{evidence-based} generative engine optimization agenda---one grounded in
measured causal weights rather than the untested heuristics currently in
circulation.

The remainder of the paper proceeds as follows.
Section~\ref{sec:background} develops the theoretical background and the
hypotheses, situating the LLM as an infomediary and deriving the reputation-signal
expectations from signaling theory and the eWOM literature.
Section~\ref{sec:method} details the audit and conjoint design, the panel of
models, the pre-specified estimands, and the analysis plan.
Section~\ref{sec:results} reports the AMCE estimates, the position-bias
trade-offs, the persona heterogeneity, and the stated-versus-revealed analysis.
Section~\ref{sec:discussion} interprets the findings for signaling theory and
algorithmic gatekeeping, sets out the implications for hotel managers, platforms,
and policy, and states the study's limitations and the research agenda it opens.

\providecommand{\hypbox}[2]{
  \begin{quote}\noindent\textbf{#1}\enspace #2\end{quote}}

\section{Background and hypotheses}\label{sec:background}

We develop the study's expectations in five steps. We first situate the large
language model (LLM) as a newly consequential infomediary in travel discovery and
identify the gap our audit fills (Section~\ref{sub:discovery}). We then derive
directional hypotheses for seven hotel reputation signals from signaling theory
and the electronic word-of-mouth (eWOM) literature (Section~\ref{sub:signals}),
specify when those signals should be reweighted---by signal diagnosticity and by
traveler persona (Section~\ref{sub:contingency})---and motivate the
position-bias and transparency research questions
(Section~\ref{sub:infomediary}). Finally, we relate the design to the closest
prior work and to the algorithm-audit tradition, and we delimit our estimand from
the silicon-sampling debate (Section~\ref{sub:priorwork}).

\subsection{AI-mediated travel discovery}\label{sub:discovery}

Travelers increasingly begin hotel discovery not at a search engine or an
online travel agency but in conversation with a general-purpose LLM assistant.
Industry tracking indicates that roughly forty percent of United States
travelers used generative-AI tools to plan a trip during 2025---an
eleven-point year-over-year increase---while the share starting trip planning at
a conventional search engine fell from about half to roughly a third over the
same period and the share starting at a generative-AI platform more than doubled
\citep{phocuswright2025}. With assistants now able to surface and, on some
platforms, transact bookings directly within the chat, the LLM increasingly
occupies the position once held by ranked search results: it decides which of
many eligible properties a traveler actually sees recommended.

This shift turns the assistant into an \emph{algorithmic infomediary} whose
selection function has direct commercial consequences for hotels. Suppliers have
already begun to respond. The emerging practice of generative engine
optimization---optimizing content for inclusion and prominence in
generative-engine answers, the successor to search engine optimization---treats
the LLM's selection behavior as an object to be influenced
\citep{aggarwal2024geo}. Yet the inputs that actually move an assistant's
recommendation are not publicly documented, and an individual hotel cannot
observe why it was or was not surfaced. The managerially urgent question is
therefore which \emph{reputation signals that a property can manage}---its guest
rating, the volume and freshness of its reviews, whether management visibly
responds, its brand affiliation, its price, and its sustainability
credentials---causally change the probability that an LLM recommends it, and with
what relative weight. Existing evidence does not answer this. The eWOM literature
quantifies how these cues move \emph{human} booking behavior and market outcomes
(Section~\ref{sub:signals}), and a nascent strand examines LLM travel
preferences (Section~\ref{sub:priorwork}), but no study has audited the
\emph{supply-side} reputation signals an LLM rewards when acting as the
gatekeeper. We close this gap with a randomized, choice-based conjoint audit that
identifies the average marginal component effect (AMCE) of each signal on the
LLM's recommendation \citep{hainmueller2014}.

\subsection{Reputation signals as quality cues}\label{sub:signals}

Hotels are experience goods: a traveler cannot verify quality before the stay.
Under information asymmetry, decision-makers infer unobservable quality from
observable cues that are informative because they are costly or difficult to
fake---the logic of signaling theory \citep{spence1973}. Online reviews and the
metadata around them are the dominant contemporary signals of hotel quality, and
a large body of human evidence establishes both their effects and, crucially,
their \emph{relative ordering}. We take that ordering as the prior against which
the LLM's revealed weights are read.

Among human-facing signals, the valence of reviews---the average guest
rating---is consistently the strongest lever. A hospitality-specific
meta-analysis reports an average eWOM \emph{valence} elasticity of hotel
performance of $0.888$, against a \emph{volume} elasticity of only $0.055$,
placing rating more than an order of magnitude above review count
\citep{yang2018}. Cross-category meta-analyses concur that valence is the
more effective cue, while confirming that volume also contributes
\citep{babicrosario2016,you2015,floyd2014}. Foundational primary studies link
review valence to sales and bookings \citep{chevalier2006,ye2009} and to booking
intentions and trust \citep{sparks2011}. Accordingly, and consistent with rating
being the headline quality signal, we expect:

\hypbox{H1.}{Average guest rating increases P(recommended).}

Review volume carries information of its own---a popularity cue and, more
importantly, an indicator of how \emph{precisely} the displayed rating is
estimated---though its standalone association with outcomes is weaker than
valence \citep{yang2018,park2015}. We therefore expect a positive but
secondary main effect:

\hypbox{H2.}{Review volume increases P(recommended).}

The recency of reviews is a freshness and relevance cue: a recent review better
reflects a property's current operation than an old one, and stale reviews are
discounted as less diagnostic of present quality. We expect recency to act as a
positive signal:

\hypbox{H3.}{Review recency (recent vs stale) increases P(recommended).}

A visible management response is a costly, observable signal of attentiveness and
service-recovery orientation. In the clearest human estimate, hotels that begin
responding to reviews experience a measurable improvement in their online
reputation, consistent with both a direct signaling effect and a shift in the
composition of subsequent reviews \citep{proserpio2017}. If an LLM has
internalized this signal, the presence of a response line should raise
recommendation probability:

\hypbox{H4.}{Visible management response increases P(recommended).}

Brand affiliation functions as a risk-reduction signal for experience goods:
chains standardize service and thereby lower the perceived variance of quality,
which is valuable when the traveler cannot inspect the property
\citep{spence1973}. We therefore expect chain affiliation to be favored over
independent status:

\hypbox{H5.}{Chain affiliation increases P(recommended) (brand-as-risk-reduction).}

Price enters as the standard economic cost of the alternative. Holding quality
signals fixed, a higher nightly rate should reduce the probability that a
property is chosen, as in conventional discrete-choice and conjoint evidence on
hotel booking \citep{assaker2023}:

\hypbox{H6.}{Price decreases P(recommended).}

Finally, third-party eco-certification is a credence signal: travelers cannot
verify a hotel's environmental practice directly, so a recognized label
substitutes. Human evidence finds that green certification does affect booking
choice, but ranks below price, rating, and other core attributes, with a modest
willingness-to-pay premium concentrated among environmentally motivated
travelers \citep{assaker2023}. We expect a positive, secondary main effect:

\hypbox{H7.}{Eco-certification increases P(recommended).}

\subsection{Signal diagnosticity and persona contingency}\label{sub:contingency}

Reputation signals are not read in isolation, and their weight should depend on
how \emph{diagnostic} each signal is and on \emph{who} the traveler is. Two
diagnosticity interactions follow from the eWOM literature. First, the value of a
high rating depends on how precisely it is estimated: a $4.7$ computed from
thousands of reviews is a more credible quality claim than the same average from
a handful, so volume should amplify the rating effect rather than merely add to
it \citep{park2015,yang2018}. Second, accumulated volume is only
informative if the underlying reviews remain current; when the most recent review
is stale, a large review count is a weaker guarantee of present quality, so
recency should moderate the volume effect. These yield:

\hypbox{H8.}{Rating $\times$ volume: the rating effect is larger at higher review volume (signal diagnosticity).}

\hypbox{H9.}{Volume $\times$ recency: the volume effect is attenuated when the most recent review is stale.}

The remaining interactions concern persona contingency. A well-calibrated
assistant should reweight cues according to the stated traveler, mirroring
documented human segment heterogeneity. Budget-constrained families are the most
price-sensitive segment, so the negative price effect should be strongest for
them. Environmentally motivated travelers carry most of the green premium, so the
eco-certification effect should be strongest for the eco-couple persona. Business
travelers most value brand consistency, loyalty standing, and the risk reduction
a chain provides, so the chain effect should be strongest for them. We therefore
expect:

\hypbox{H10.}{Persona $\times$ price: the (negative) price effect is strongest for the budget-family persona.}

\hypbox{H11.}{Persona $\times$ sustainability: the eco-certification effect is strongest for the eco-couple persona.}

\hypbox{H12.}{Persona $\times$ affiliation: the chain effect is strongest for the business persona.}

Hypotheses H1--H12 constitute the confirmatory family, estimated on the pooled
panel of audited models with model fixed effects and corrected for multiple
comparisons. We emphasize that the contribution does not depend on the direction
of any single effect: a null or sign-reversed result---reported with an
equivalence test---would itself be a substantive finding about how the
gatekeeper's reputation weights diverge from the documented human ordering
summarized in Section~\ref{sub:signals}.

\subsection{The algorithmic infomediary: position and transparency}\label{sub:infomediary}

Beyond the content of the cards, the assistant's own interface introduces a
potential distortion that no hotel controls: the order in which candidate
properties are presented. Computer-science evaluations consistently find that LLM
recommenders and zero-shot rankers are highly sensitive to the order of
candidate items, such that reordering otherwise identical options changes the
output \citep{bito2025,hou2024}, and related work documents systematic
identity- and selection-order biases in LLM choice among candidates
\citep{ren2024}. This prior work establishes that position bias
\emph{exists} and proposes prompting strategies to mitigate it, but it has not
causally quantified the magnitude of the effect in a realistic travel decision,
nor expressed it in units a manager or platform can interpret. Because our design
randomizes list position independently of card content, the position AMCE is
causally identified and can be converted into a price-equivalent---how many
dollars per night, or how many rating steps, a more prominent slot is worth. We
treat this as a research question rather than a directional hypothesis:

\hypbox{RQ-Position.}{Does list position causally affect recommendation probability, and what is its price-equivalent magnitude?}

A second question concerns transparency. Each model is asked not only to choose
but to state a reason, allowing us to compare the attributes a model
\emph{names} with the attributes that causally drive its choices, as recovered
from the AMCEs. A mismatch between stated and revealed weights would bear on the
interpretability of these systems as accountable intermediaries
\citep{metaxa2021}. We again pose this as an exploratory, descriptive question:

\hypbox{RQ-Reasons.}{Do models' stated reasons correspond to their revealed attribute weights?}

\subsection{Relation to prior work}\label{sub:priorwork}

Our study sits at the intersection of three literatures and is differentiated from
the closest work in each. Methodologically, it extends the
\emph{algorithm-audit} tradition. Correspondence audits established the template
of randomizing a unit's displayed attributes to measure differential treatment
causally \citep{bertrand2004}; the algorithm-audit program adapted this logic to
automated systems and catalogued its designs \citep{sandvig2014,metaxa2021}. We
implement an audit of an LLM in which a fully randomized, choice-based conjoint
supplies identification, recovering AMCEs as the primary estimand
\citep{hainmueller2014} with conditional-logit and price-equivalent analyses as
secondary specifications \citep{mcfadden1974,krinsky1986}; scale-free comparison
of effects across models and decoding temperatures is justified by the role of
the scale parameter in choice models \citep{swait1993}.

Substantively, the closest prior work is \citet{reusens2026}, who present LLMs
with travel choice dilemmas, fit multinomial-logit models to infer the
implied willingness to pay for room attributes such as a view, and compare these
to human economic benchmarks, finding meaningful estimates for larger models
alongside systematic attribute-level deviations. That study and ours are
complementary but distinct in unit, theory, and contribution. \citet{reusens2026}
ask a \emph{demand-side} question---does the LLM value room attributes as a
consumer would?---and recover preference parameters. We ask a \emph{supply-side}
question---which reputation signals that a hotel can actively manage move the
machine's recommendation, and how does platform-controlled position intrude?---and
frame it through algorithmic gatekeeping of property-level visibility, with a
hospitality-management contribution and direct relevance to generative engine
optimization. Their finding that even capable models deviate from human valuations
at the attribute level directly motivates our concern that the gatekeeper's
reputation weights may diverge from the well-established human eWOM ordering.

A potential objection invokes the active debate over using LLMs as synthetic
respondents. ``Silicon sampling'' has shown that persona-conditioned models can
reproduce some human subgroup patterns \citep{argyle2023,horton2023}, but
critics document severe failures---persona conditioning can amplify stereotypes
and inflate group differences several-fold relative to human benchmarks
\citep{bisbee2024}---prompting formal cautions against treating LLM output as a
stand-in for human populations \citep{sarstedt2024}. These critiques do not
threaten our design; they motivate it. We do not use the LLM to estimate human
preferences, and we make no claim that our AMCEs describe human travelers. Our
estimand is the behavior of the LLM \emph{as a deployed gatekeeper}, because the
assistant is the system travelers actually consult when they ask it to recommend
a hotel. The very distortions the silicon-sampling literature warns about---
fidelity gaps, stereotype amplification, and the collapse of response
variance---are precisely the phenomena an audit of a consequential intermediary
should detect and quantify. We compare the LLM's weights to human eWOM benchmarks
not to validate the model as a human proxy, but to characterize how the
gatekeeper's revealed reputation function \emph{diverges} from the human evidence
that hotels have long been managed against.

\section{Method}\label{sec:method}

\subsection{Design overview and rationale}\label{sec:method-overview}

We treat the language model as an algorithmic infomediary and study it with an
algorithm audit: a controlled experiment in which inputs are randomized and the
system's outputs are observed to recover the input--output mapping the system
enforces in practice \citep{sandvig2014,metaxa2021}. Audit studies descend from
the correspondence tradition, in which paired applications differing only in a
randomized characteristic reveal differential treatment \citep{bertrand2004};
algorithm audits port this logic to automated decision-makers. Because the
profiles are synthetic and the manipulated attributes are assigned at random,
the design identifies the \emph{causal} effect of each attribute on the system's
output---which observational analyses of scraped recommendations cannot, as
there attributes are confounded with unobserved demand, quality, and listing
strategy.

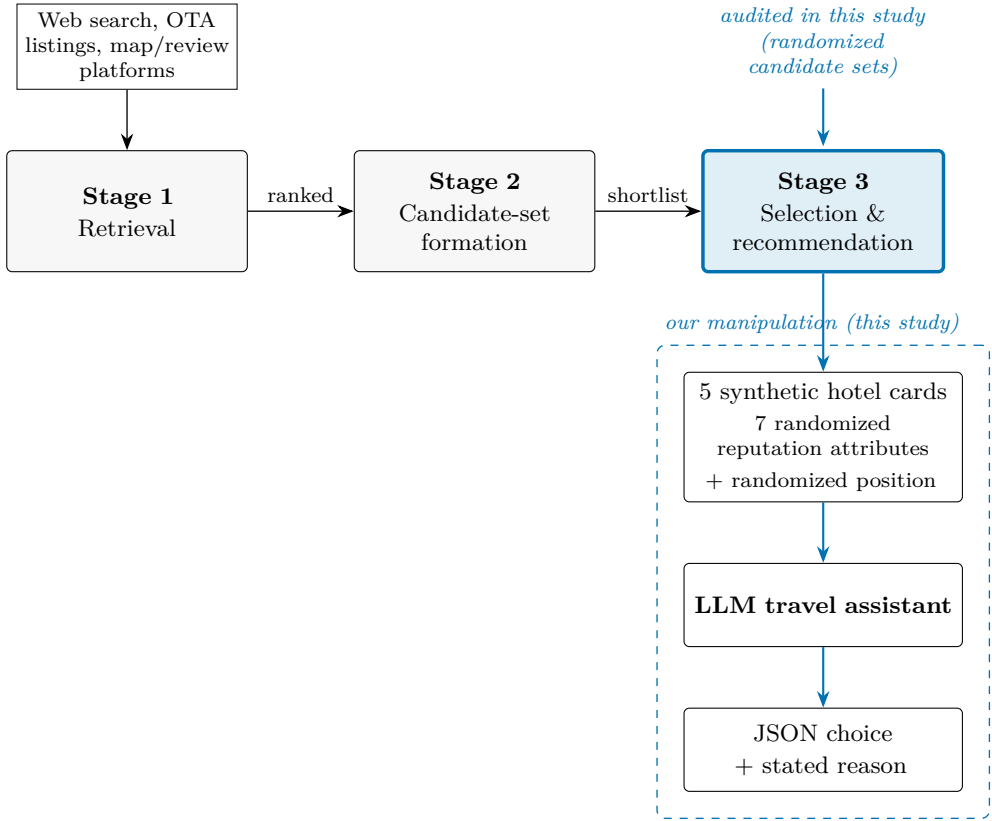
\begin{figure}[t]
\centering
{
\definecolor{auditblue}{RGB}{0,114,178}
\definecolor{auditorange}{RGB}{230,159,0}
\begin{tikzpicture}[
    font=\footnotesize,
    >={Stealth[length=2.2mm]},
    stage/.style={
      draw, rounded corners=2pt, line width=0.4pt,
      align=center, inner sep=4pt,
      minimum height=16mm, text width=29mm, fill=black!3},
    audited/.style={
      draw, rounded corners=2pt, line width=1.3pt, draw=auditblue,
      align=center, inner sep=4pt,
      minimum height=16mm, text width=29mm, fill=auditblue!12},
    src/.style={
      draw, line width=0.3pt, align=center, inner sep=3pt,
      font=\scriptsize, text width=27mm, minimum height=5mm, fill=white},
    manip/.style={
      draw, rounded corners=2pt, line width=0.4pt,
      align=center, inner sep=4pt, minimum height=11mm,
      text width=34mm, fill=white},
    flow/.style={->, line width=0.5pt},
    flowb/.style={->, line width=0.8pt, draw=auditblue},
  ]

  \node[stage] (s1) {\textbf{Stage 1}\\[1pt]Retrieval};
  \node[stage, right=14mm of s1] (s2) {\textbf{Stage 2}\\[1pt]Candidate-set formation};
  \node[audited, right=14mm of s2] (s3) {\textbf{Stage 3}\\[1pt]Selection \&\\recommendation};

  \node[src, above=8mm of s1] (websrc) {Web search, OTA listings, map/review platforms};
  \draw[flow] (websrc) -- (s1);

  \draw[flow] (s1) -- node[above=-0.5pt, font=\scriptsize]{ranked} (s2);
  \draw[flow] (s2) -- node[above=-0.5pt, font=\scriptsize]{shortlist} (s3);

  \node[align=center, text width=34mm, font=\scriptsize\itshape,
        above=8mm of s3, text=auditblue] (annot)
        {audited in this study\\(randomized candidate sets)};
  \draw[->, line width=0.8pt, draw=auditblue, shorten >=1pt]
        (annot.south) -- (s3.north);

  \node[manip, below=13mm of s3] (cards)
        {\hyphenpenalty=10000\exhyphenpenalty=10000
         5 synthetic hotel cards\\[2pt]\scriptsize 7 randomized
         reputation attributes\\[1pt]$+$ randomized position};
  \node[manip, below=8mm of cards] (llm)
        {\textbf{LLM travel assistant}};
  \node[manip, below=8mm of llm] (out)
        {JSON choice\\[1pt]$+$ stated reason};

  \draw[flowb] (s3.south) -- (cards.north);
  \draw[flowb] (cards) -- (llm);
  \draw[flowb] (llm) -- (out);

  \node[draw=auditblue, dashed, line width=0.5pt, rounded corners=3pt,
        fit=(cards)(llm)(out), inner sep=3.5mm] (mbox) {};
  \node[anchor=south west, font=\scriptsize\itshape, text=auditblue,
        inner sep=0pt] at ([xshift=1mm,yshift=1.3mm]mbox.north west)
        {our manipulation (this study)};

\end{tikzpicture}
}

\caption{LLM-mediated hotel discovery and the audited stage. The assistant's
pipeline is decomposed into retrieval, candidate-set assembly, and selection;
this study audits the selection stage by presenting randomized candidate sets
under experimental control.}
\label{fig:pipeline}
\end{figure}

An LLM-assisted booking pipeline can be decomposed into three stages: retrieval
of candidate properties from an index or tool call, assembly of a candidate set,
and selection of a recommendation from that set
(Figure~\ref{fig:pipeline}). We audit the final,
\emph{selection} stage in isolation. Holding the candidate set fixed and
exogenous and varying only the reputation signals attached to its members
isolates the model's selection rule from the confounding influence of retrieval,
which is itself opaque, provider-specific, and prone to drift. This mirrors the
logic of choice-based conjoint analysis, in which respondents repeatedly choose
from sets of multi-attribute profiles and the analyst recovers attribute weights
from the choices \citep{hainmueller2014}. The model occupies the role of the
deciding agent; the randomized profiles play the role of the conjoint
alternatives.

Concretely, the experiment is a randomized, choice-based conjoint task. On each
trial the model receives a traveler request and a set of five hotel
alternatives, each described by seven randomly assigned reputation attributes,
and is asked to recommend exactly one property. Repeating this over thousands of
trials with independently randomized attributes lets us estimate, for each
attribute, the average change in the probability that a property is recommended
when that attribute is moved across its levels. The full protocol, hypotheses,
and analysis decisions were specified before any main-arm data were
collected, and the experimental design (the choice sets and their attribute
assignments) was generated from a fixed master seed and cryptographically
hashed in advance, so that the design provably precedes the runs
(Section~\ref{sec:method-transparency}).

\subsection{Stimuli}\label{sec:method-stimuli}

Each hotel alternative was rendered as a short property card carrying seven
manipulated attributes drawn from the reputation and electronic word-of-mouth
literature: the displayed average guest rating, the number of reviews on which
that rating is based, the recency of the most recent review, whether management
is shown to respond to reviews, chain versus independent affiliation, nightly
price, and the presence of a third-party eco-certification (Green Key).
Table~\ref{tab:attributes} lists the attributes, their levels, and how each was
operationalized in the card. Attribute levels were assigned independently and
with equal probability to each of the five alternatives in a choice set, subject
to a no-duplicate rule (no two alternatives within a set shared an identical
profile across all seven attributes), so that attributes are mutually orthogonal
by construction and orthogonal to display position in expectation.

\begin{table}[t]
\centering
\caption{Manipulated hotel attributes, levels, and card operationalization.}
\label{tab:attributes}
\begin{threeparttable}
\begin{tabular}{@{}l l p{5.7cm}@{}}
\toprule
Attribute & Levels & Operationalization in the card \\
\midrule
Guest rating & 3.9, 4.3, 4.7 & ``Guest rating: \textit{r}/5'' \\
Review volume & 45, 420, 2{,}100 & ``(\textit{v} reviews)'' on the rating line \\
Review recency & 3 days / 11 months & ``Most recent review: \textit{recency}'' \\
Mgmt.\ response & present, absent & ``Management responds to guest reviews''\tnote{a} \\
Affiliation & chain, independent & ``Part of a major international hotel chain'' vs.\ ``Independently operated hotel'' \\
Price (USD/night) & 129, 189, 249 & ``Price: \$\textit{p} per night'' \\
Sustainability & present, absent & ``Certified eco-friendly (Green Key)''\tnote{a} \\
\bottomrule
\end{tabular}
\begin{tablenotes}\footnotesize
\item[a] Binary attributes were operationalized by the presence or absence of the
line; no negated ``does not respond'' / ``not certified'' text was shown.
\end{tablenotes}
\end{threeparttable}
\end{table}

Two design elements were held constant across every alternative as stated
controls: each property was described as a ``4-star hotel'' and as ``Located
near the city center.'' Fixing star class and location removes two salient
quality and convenience cues that would otherwise compete with the reputation
signals of interest, so that any differential treatment is attributable to the
manipulated attributes rather than to category or distance. Hotel names were
drawn without replacement, per choice set, from a fixed pool of 40 neutral
surname-style labels (e.g., ``Hotel Arden,'' ``Hotel Bellmore,'' ``Hotel
Calloway''), chosen to be free of obvious national, gendered, or quality
connotations; names were assigned to alternatives independently of their
attributes, which supports the pre-specified name-placebo test
(Section~\ref{sec:method-analysis}). Because names are randomized with respect to
attributes, the pool functions as a counterbalanced set of arbitrary identifiers
rather than as a treatment.

A representative card, as presented to the model, reads:

\begin{quote}\ttfamily
Hotel Marlowe - 4-star hotel\\
Located near the city center\\
Guest rating: 4.7/5 (2100 reviews)\\
Most recent review: 3 days ago\\
Price: \$189 per night\\
Part of a major international hotel chain\\
Management responds to guest reviews\\
Certified eco-friendly (Green Key)
\end{quote}

The five cards for a trial were concatenated in randomized display order and
embedded in a prompt template (Section~\ref{sec:method-personas}). Two
presentation formats were used. In the primary \emph{card} format, attributes
appear as the labeled lines shown above. In a secondary \emph{grounded-snippet}
format, the identical attribute values are re-expressed as a retrieved
web-search result attributed to a fictitious review site (e.g., ``[stayreviews.com]
Hotel Marlowe $\mid$ 4-star hotel, located near the city center. Rated 4.7/5
across 2100 guest reviews\ldots''), which probes whether the selection rule is
sensitive to the surface form in which the same evidence is grounded. The two
formats convey the same attribute information; they differ only in framing.

\subsection{Personas and prompt templates}\label{sec:method-personas}

Each prompt opened with a first-person traveler request instantiating one of
three personas, quoted verbatim from the configuration:

\begin{itemize}
\item \emph{Business:} ``I am traveling alone for work. I have meetings during
the day and value convenience and reliability.''
\item \emph{Family:} ``We are a family of four traveling with two young
children, and we are watching our budget on this trip.''
\item \emph{Eco-couple:} ``My partner and I are taking a short city break.
Sustainability genuinely matters to us when we choose where to stay.''
\end{itemize}

The personas were written to create directional, pre-specified expectations
about which attributes should matter most (price for the budget-conscious
family, sustainability for the eco-couple, brand reliability for the business
traveler), enabling the persona-by-attribute interaction tests H10--H12.

To prevent any single phrasing from driving results, the persona sentence was
embedded in one of nine paraphrased prompt templates that vary the framing of
the assistant role and the request while holding the task constant (for example,
``You are a helpful travel assistant. \{persona\} Below are the hotels available
in my destination city for my dates.\ldots Which one should I book?''). Trials
were allocated by stratified randomization over the full $3 \times 9 = 27$
persona-by-template strata, with 112 trials per stratum, yielding the
$27 \times 112 = 3{,}024$ trials of the main arm; execution order was then
shuffled so that strata are interleaved during data collection.

The model was instructed to return a single recommendation in a fixed,
choice-first JSON format. The instruction read, verbatim:

\begin{quote}\ttfamily\raggedright
Respond with ONLY a JSON object, nothing else:\\
\{"choice": "<exact hotel name>", "reason": "<one or two sentences>"\}
\end{quote}

Eliciting the choice as the first JSON field, before the free-text reason,
captures the recommendation prior to any post-hoc rationalization; a secondary
field-order arm reverses the two fields to test whether requiring the reason
first alters the choice (Section~\ref{sec:method-arms}). The free-text reason is
the basis for the stated-versus-revealed analysis (RQ4).

\subsection{Model panel}\label{sec:method-models}

The audited panel comprises twelve instruction-tuned models. Four are
open-weight systems---\texttt{llama3.2:3b}, \texttt{qwen2.5:3b},
\texttt{phi3:mini}, and \texttt{mixtral-8x7b-instruct}---served through
open-weight runtimes (Ollama for the locally hosted models), representing the
quantized, deployable assistants increasingly embedded in consumer travel and
search applications; the exact build identifier returned by the runtime is
logged with every call. Eight are proprietary hosted-API models accessed through
an OpenAI-compatible interface: OpenAI GPT-4o-mini, three Google Gemini models
(1.5 Pro, 2.0 Flash, 2.0 Pro), and four Anthropic Claude models. In keeping with
the pre-specified protocol, the exact API model identifiers were verified against
each provider's model endpoint and recorded at run start. Reporting the same audit
across open-weight and proprietary panels separates idiosyncrasies of a single
model from regularities of the selection rule that recur across the model class.

Unless a robustness arm specifies otherwise, generation used a sampling
temperature of 0.7 with nucleus sampling at $\text{top-}p = 0.9$ and a cap of
220 output tokens. A nonzero temperature is the realistic deployment setting for
a consumer-facing assistant and also aligns the data-generating process with the
random-utility interpretation of the choice models, under which stochastic
choice reflects an unobserved utility shock rather than measurement error; a
deterministic temperature-0 arm is included as a robustness condition
(Section~\ref{sec:method-arms}). Every call drew a deterministic per-call seed
equal to \texttt{md5(model\,$\mid$\,arm\,$\mid$\,trial\,$\mid$\,rep)}, so that
runs are reproducible and resume-safe while remaining independent across cells.
The complete generation configuration (temperature, top-$p$, token cap,
retries, and timeout) is logged alongside the model identifier for every call.

\subsection{Experimental arms and procedure}\label{sec:method-arms}

\begin{figure}[t]
\centering
{
\definecolor{auditblue}{RGB}{0,114,178}
\definecolor{auditorange}{RGB}{230,159,0}
\begin{tikzpicture}[
    font=\footnotesize,
    >={Stealth[length=2.4mm]},
    band/.style={
      draw, rounded corners=2pt, line width=0.4pt, align=center,
      inner sep=4pt, fill=black!3, text width=126mm},
    design/.style={band, line width=0.9pt, draw=auditblue,
      fill=auditblue!10, font=\footnotesize},
    exec/.style={band, fill=black!5},
    analysis/.style={band, fill=auditblue!8, draw=auditblue, line width=0.5pt},
    sbox/.style={
      draw, rounded corners=2pt, line width=0.4pt, align=center,
      inner sep=3pt, fill=black!3, font=\scriptsize, text width=33mm,
      minimum height=10mm},
    pbox/.style={
      draw, rounded corners=2pt, line width=0.4pt, align=center,
      inner sep=3pt, fill=black!3, font=\scriptsize, text width=49mm,
      minimum height=9mm},
    arm/.style={
      draw, rounded corners=1.5pt, line width=0.3pt, align=center,
      inner sep=1.5pt, fill=auditorange!12, font=\scriptsize,
      minimum height=12mm, text width=12.5mm},
    armmain/.style={arm, line width=0.7pt, draw=auditorange,
      fill=auditorange!22},
    tag/.style={font=\scriptsize\bfseries, text=black!55,
      anchor=south east, inner sep=0pt},
    flow/.style={->, line width=0.7pt, draw=black!70},
  ]

  \node[design] (design)
    {\textbf{Master design:} 3{,}024 choice sets (seed 20260612,
     SHA-256 hashed),\\ stratified over 3 personas $\times$ 9 templates};

  \node[sbox, below=9mm of design.south, anchor=north, xshift=-39.5mm] (stimC)
    {Card rendering};
  \node[sbox, right=5mm of stimC] (stimS) {Grounded-snippet rendering};
  \node[sbox, right=5mm of stimS] (stimR) {Reversed-layout rendering};

  \node[armmain, below=10mm of stimC.south, anchor=north, xshift=-18.5mm] (a1)
    {main\\\textbf{3{,}024}\\$T{=}0.7$};
  \node[arm, right=1mm of a1] (a2) {temp-0\\\textbf{594}};
  \node[arm, right=1mm of a2] (a3) {grounded\\\textbf{297}};
  \node[arm, right=1mm of a3] (a4) {rank-3\\\textbf{297}};
  \node[arm, right=1mm of a4] (a5) {field-\\order\\\textbf{189}};
  \node[arm, right=1mm of a5] (a6) {retest\\\textbf{108}\\$\times5$};
  \node[arm, right=1mm of a6] (a7) {letter\\\textbf{297}};
  \node[arm, right=1mm of a7] (a8) {no-\\persona\\\textbf{297}};
  \node[arm, right=1mm of a8] (a9) {reversed\\\textbf{189}};

  \node[pbox, below=10mm of a5.south, anchor=north, xshift=-27.5mm] (panelL)
    {4 local open-weight models\\(Ollama, quantized)};
  \node[pbox, right=6mm of panelL] (panelF)
    {8 API models\\(OpenAI, Google, Anthropic)};

  \node[exec, below=10mm of panelL.south, anchor=north, xshift=27.5mm] (exec)
    {\textbf{Checkpointed runner:} JSON parse $\rightarrow$ 2 retries
     $\rightarrow$ fuzzy name match; per-call seeds};

  \node[analysis, below=9mm of exec.south, anchor=north] (analysis)
    {\textbf{Analysis:} AMCEs (trial-clustered) $\mid$ conditional logit
     $\mid$ price-equivalents (Krinsky--Robb)\\
     stated-vs-revealed $\mid$ robustness battery};

  \node[tag] at ([yshift=1.5pt]design.north west) {(a)};
  \node[tag] at ([yshift=1.5pt]stimC.north west) {(b)};
  \node[tag] at ([yshift=1.5pt]a1.north west) {(c)};
  \node[tag] at ([yshift=1.5pt]panelL.north west) {(d)};
  \node[tag] at ([yshift=1.5pt]exec.north west) {(e)};
  \node[tag] at ([yshift=1.5pt]analysis.north west) {(f)};

  \coordinate (spine) at ($(panelL.east)!0.5!(panelF.west)$);
  \draw[flow] (design.south) -- (design.south |- stimS.north);
  \draw[flow] (stimS.south) -- (stimS.south |- a5.north);
  \draw[flow] (a5.south) -- (a5.south |- panelL.north);
  \draw[flow] (spine |- panelL.south) -- (spine |- exec.north);
  \draw[flow] (exec.south) -- (analysis.north);

\end{tikzpicture}
}

\caption{Experiment architecture. The hashed master design is rendered into
stimuli, fielded across experimental arms and the model panel by a checkpointed
runner, and analyzed under the pre-specified statistical analysis plan.}
\label{fig:architecture}
\end{figure}
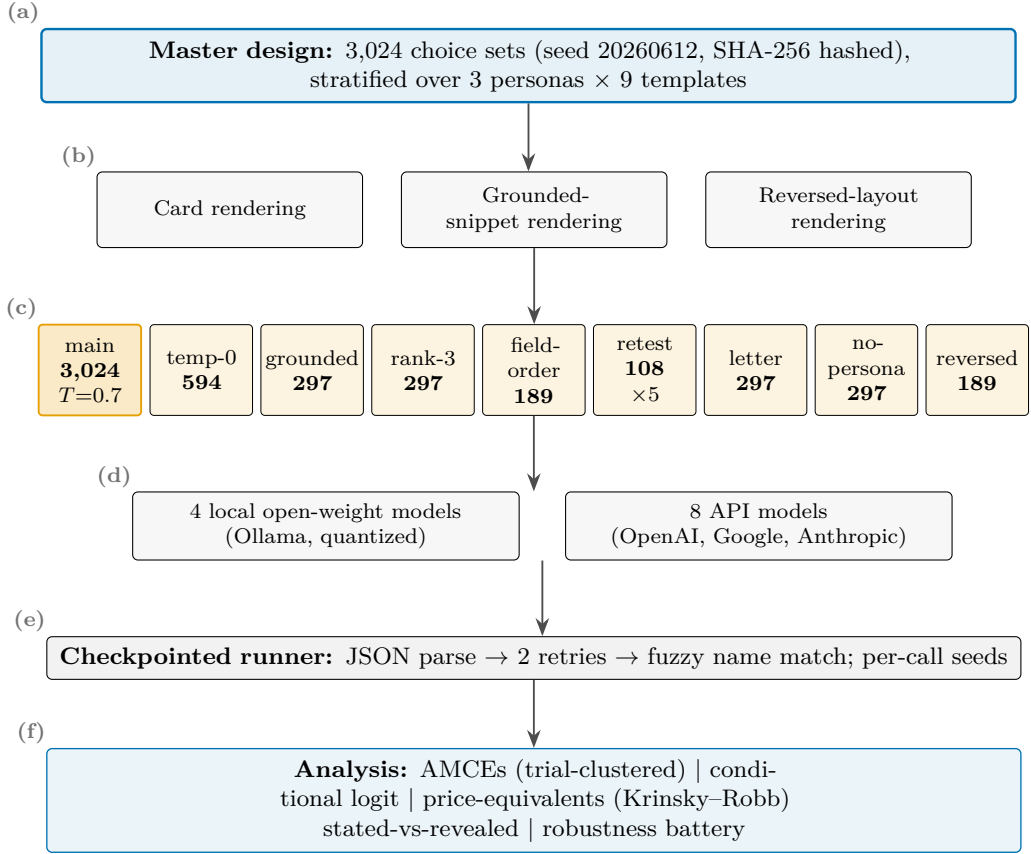

The main arm comprises the 3,024 trials described above. To probe the robustness
of the selection rule and to address anticipated reviewer concerns about prompt
and elicitation construction, we additionally fielded a set of secondary arms,
each a disjoint, stratified subset of the main design that re-presents
\emph{identical stimuli} under a single changed condition (Table~\ref{tab:arms}).
Figure~\ref{fig:architecture} summarizes the full experimental architecture from
design generation through execution to analysis.

Holding the choice sets fixed across arms means that any difference in estimated
attribute weights is attributable to the manipulated condition (temperature,
format, elicitation, persona text, or card layout) rather than to differences in
the underlying profiles. Because all secondary arms are carved from the main
design, they consume no new choice sets and inherit its stratification (an
integer number of trials per persona-by-template stratum in every arm). The
\texttt{nopersona} and \texttt{card\_reversed} arms were added in a
pre-data amendment to the analysis plan and are preplanned exploratory
robustness checks; they are not part of the confirmatory family.

\begin{table}[t]
\centering
\caption{Experimental arms. All arms share common stimuli within each arm across
the model panel; every secondary arm is a disjoint, stratified subset of the
main design re-presenting identical choice sets under one changed condition.}
\label{tab:arms}
\begin{threeparttable}
\begin{tabular}{@{}llll@{}}
\toprule
Arm & Trials & Condition changed vs.\ main & Temp. \\
\midrule
\texttt{main}          & 3{,}024 & --- (reference) & 0.7 \\
\texttt{temp0}         & 594 & deterministic decoding & 0.0 \\
\texttt{grounded}      & 297 & grounded web-snippet format & 0.7 \\
\texttt{rank\_top3}    & 297 & elicit a top-3 ranking & 0.7 \\
\texttt{field\_order}  & 189 & reason elicited before choice & 0.7 \\
\texttt{retest}        & 108\tnote{a} & repeated under distinct seeds & 0.7 \\
\texttt{letter}        & 297 & letter-labeled choice (logprobs) & 0.7 \\
\texttt{nopersona}     & 297 & persona replaced by neutral text & 0.7 \\
\texttt{card\_reversed}& 189 & card attribute lines reversed & 0.7 \\
\addlinespace
\texttt{manip}         & 20\tnote{b} & manipulation check (per model) & 0.7 \\
\bottomrule
\end{tabular}
\begin{tablenotes}\footnotesize
\item[a] 108 distinct choice sets, each re-presented under 5 distinct seeds
(540 calls per model); retest repetitions are excluded from the main estimation.
\item[b] 20 free-text calls per model asking what Green Key certification is, to
confirm the models can interpret the least familiar attribute; not a choice task.
\end{tablenotes}
\end{threeparttable}
\end{table}

Table~\ref{tab:arms} details each secondary arm. Briefly: \texttt{temp0}
isolates the selection rule from sampling noise; \texttt{grounded} and
\texttt{card\_reversed} vary the surface form and the attribute ordering of the
cards (the latter testing whether the rating-first layout inflates the rating
weight); \texttt{field\_order} reverses the JSON fields; \texttt{rank\_top3}
elicits a ranked top-three (analyzed by rank-ordered logit); \texttt{letter}
labels alternatives A--E independently of display position to enable a
token-logprob decomposition where available; \texttt{nopersona} replaces the
persona with neutral text as a persona-free baseline; and \texttt{retest}
re-presents 108 sets under five seeds each to quantify within-item response
stability.

Before the main runs, a 50-trial pilot per model (the leading slice of the main
arm) was used only to verify JSON parse rates and timing; pilot calls are stored
in separate checkpoint files and are never pooled with the analyzed data. During
data collection, a call whose output failed JSON parsing was re-asked up to two
times with a stricter formatting reminder; calls still unparseable after two
retries are recorded as parse failures and excluded from the choice models.
Parse failures are not assumed ignorable: missingness is modeled explicitly by
regressing failure on attributes, persona, template, and model, and if any model
exceeds a 5\% failure rate, worst-case bounds are reported and that model is
flagged. All inference calls are checkpointed to disk and resume-safe.

\subsection{Statistical analysis}\label{sec:method-analysis}

The primary estimand is the average marginal component effect (AMCE) of each
attribute, expressed in percentage points of the probability that an alternative
is chosen \citep{hainmueller2014}. We estimate AMCEs from a linear probability
model in which the binary choice indicator is regressed on the attribute levels
(rating, review volume, and price entered as ordered categorical dummies against
their lowest level; the binary attributes as 0/1 indicators) together with
display-position dummies; the pooled specification adds model fixed effects.
Standard errors are clustered by trial (choice set), which accounts for the
dependence among the five alternatives within a set and, in the pooled model,
for the dependence induced by presenting common stimuli to every model. The AMCE
is the headline quantity precisely because it is scale-free: discrete-choice
coefficients are identified only up to a model-specific scale (the inverse of
the logit error variance), so raw coefficients are not comparable across models
or across temperatures, whereas choice-probability effects are
\citep{swait1993}.

The confirmatory family is exactly the twelve pre-specified hypotheses H1--H12,
comprising the seven attribute main effects (each a joint Wald test over the
attribute's levels on the pooled model) and five prespecified interactions
(rating $\times$ volume, volume $\times$ recency, and the three
persona-by-attribute terms), added in a second specification. Familywise error
across these twelve tests is controlled by the Holm procedure at
$\alpha = .05$. All other estimates---per-model AMCEs, the position effect, the
stated-versus-revealed comparison, and all arm contrasts---are reported as
preplanned but exploratory, without a familywise claim. We also report each
attribute's importance share (the range of its AMCEs divided by the summed
ranges across attributes).

As a secondary, structurally explicit analysis we fit conditional
(multinomial) logit choice models grouped by trial, both per model and pooled
with model fixed effects and model-by-attribute interactions
\citep{mcfadden1974}; coefficients and odds ratios are reported alongside the
AMCEs. In the deterministic \texttt{temp0} arm we anticipate quasi-separation
and accordingly report penalized (Firth) estimates or choice shares and AMCEs
only; the \texttt{rank\_top3} arm is analyzed with a rank-ordered (exploded)
logit.

Trade-offs between non-price attributes and price are computed only if a
pre-specified monotonicity gate is satisfied---namely that the price AMCEs
decrease monotonically across \$129, \$189, and \$249. When the gate holds, each
attribute's effect is converted to a price-equivalent by dividing its AMCE by
the per-dollar price AMCE linearized over the \$120 price range, per persona and
pooled, with 95\% confidence intervals obtained by the Krinsky--Robb parametric
bootstrap (10,000 draws from the estimated coefficient distribution)
\citep{krinsky1986}. The headline trade-off is the position-1 advantage
expressed in dollars per night and in rating-step equivalents. Any claim that an
attribute has \emph{no} effect is supported not by a non-significant test but by
two one-sided tests (TOST) against a smallest effect size of interest of
$|\text{AMCE}| < 1.5$ percentage points. The name-placebo expectation (that the
arbitrary hotel names carry no weight) is evaluated with a joint likelihood-ratio
test over the hotel-name dummies, which is expected to be null.

The design's precision was established before data collection by Monte-Carlo
simulation under the pre-specified estimator. Table~\ref{tab:power} reports, for
each effect, the minimum detectable effect (MDE) at 80\% power given the
simulated sampling variability of the main arm. The MDEs for the seven attribute
main effects and the position effect span roughly 1.8--2.9 percentage points,
below all effect sizes we regard as substantively meaningful, so the main arm is
well powered for the confirmatory family; the prespecified rating $\times$ volume
interaction has a larger MDE, as expected for an interaction term.

\begin{table}[t]
\centering
\caption{Monte-Carlo minimum detectable effects (MDE) at 80\% power for the
main arm, by effect, in percentage points of choice probability. Values are
reproduced from \texttt{results/tables/power\_mde.csv}.}
\label{tab:power}
\begin{threeparttable}
\begin{tabular}{@{}lc@{}}
\toprule
Effect & MDE at 80\% power (pp) \\
\midrule
Guest rating         & 2.26 \\
Review volume        & 2.18 \\
Review recency       & 1.77 \\
Management response   & 1.79 \\
Affiliation          & 1.79 \\
Price                & 2.15 \\
Sustainability       & 1.86 \\
Position~1 (vs.\ later slots) & 2.87 \\
Rating $\times$ volume        & 5.36 \\
\bottomrule
\end{tabular}
\begin{tablenotes}\footnotesize
\item Estimated by simulation under the pre-specified linear-probability
estimator with trial-clustered standard errors; see \texttt{src/power\_sim.py}.
\end{tablenotes}
\end{threeparttable}
\end{table}

\subsection{Transparency and research ethics}\label{sec:method-transparency}

The study's design and analysis plan were specified, recorded, and
cryptographically hashed before any main-arm data were collected, and the
single pre-data amendment (adding the \texttt{nopersona} and
\texttt{card\_reversed} robustness arms) is documented and dated. The
experimental design was generated from a fixed master seed and recorded with
SHA-256 hashes of the design files in advance of all runs, so that the design
provably precedes data collection; the amendment was verified programmatically
to leave every pre-existing arm's per-stratum allocation bit-for-bit unchanged.
To support full reproducibility we will archive, in a public repository, the
design files and their hashes, the random seeds, the complete set of prompt
templates and personas, the generation configuration, the analysis code, and the
raw verbatim model outputs, so that every statistic in the manuscript can be
traced to a script-produced artifact rather than a hand-entered value.

The study involves no human subjects: all traveler profiles and hotel
alternatives are synthetic, and the experimental units are model responses, so
the work is exempt from human-subjects review. The only ethical consideration
specific to the materials is that the named review sites in the grounded-snippet
format are fictitious and were chosen so as not to correspond to real
businesses. Our use of generative models is confined to the audited systems and
the (non-audited) reason-coding judge; this is declared in full in accordance
with the journal's policy on generative-AI use, with the relevant statement
provided in the submission's declaration section.

\section{Results}\label{sec:results}

We report results in the order of the analysis plan. Section~\ref{sub:diag}
establishes data quality and the handling of non-responses. Section~\ref{sub:amce}
presents the primary causal estimates and the confirmatory tests of H1--H7 and
ranks the reputation signals by importance. Section~\ref{sub:interactions} reports
the diagnosticity and persona interactions (H8--H12). Section~\ref{sub:position}
isolates list-position bias and Section~\ref{sub:tradeoffs} converts effects into
price-equivalents. Section~\ref{sub:statedrevealed} compares the models' stated
reasons with their revealed weights, and Section~\ref{sub:robust} summarizes the
pre-specified robustness battery. All estimates are average marginal component
effects (AMCEs) on the probability of being recommended, in percentage points,
with standard errors clustered by choice set; confirmatory tests are
Holm-corrected over the H1--H12 family.

\subsection{Diagnostics and parse rates}\label{sub:diag}

Table~\ref{tab:parse} reports the response parse-failure rate by model and arm.
Across all 61,459 model calls the overall parse-success rate was 99.98\% (15
unparseable responses in total), and no model's aggregate failure rate exceeded
the pre-specified 5\% threshold: per-model failure rates ranged from 0.00\% to
0.26\%. A single secondary arm---the rank-ordered top-3 task for
\texttt{phi3:mini}---reached 5.1\% (15 of 297), the only cell at or above the
threshold; because it lies outside the main conjoint and the affected responses
are excluded from estimation, it does not bear on the primary estimates. With
only 15 failures in total the pre-specified missingness model could not be
fit---non-response is too rare to associate with any attribute, persona,
template, or model---so failures are treated as ignorable and no worst-case
bounds were required. The realized design is balanced: every attribute level
appears with frequency within 0.01 of its uniform target (three-level
attributes at $\approx0.333$ each, binary attributes at $\approx0.500$),
confirming that randomization supports the AMCE interpretation
(Table~\ref{tab:parse} and the balance diagnostics).

\begin{table}[t]
\centering
\caption{Response parse-failure rates.}
\label{tab:parse}
\resizebox{\textwidth}{!}{
\begin{tabular}{lrrrrrrrrr}
\toprule
Model & card\_reversed & field\_order & grounded & letter & main & nopersona & rank\_top3 & retest & temp0 \\
\midrule
claude-fable-5 & 0.0 & 0.0 & 0.0 & 0.0 & 0.0 & 0.0 & 0.0 & 0.0 & 0.0 \\
claude-haiku-4-5 & 0.0 & 0.0 & 0.0 & 0.0 & 0.0 & 0.0 & 0.0 & 0.0 & 0.0 \\
claude-opus-4-8 & 0.0 & 0.0 & 0.0 & 0.0 & 0.0 & 0.0 & 0.0 & 0.0 & 0.0 \\
claude-sonnet-4-6 & 0.0 & 0.0 & 0.0 & 0.0 & 0.0 & 0.0 & 0.0 & 0.0 & 0.0 \\
gemini-1.5-pro & 0.0 & 0.0 & 0.0 & 0.0 & 0.0 & 0.0 & 0.0 & 0.0 & 0.0 \\
gemini-2.0-flash & 0.0 & 0.0 & 0.0 & 0.0 & 0.0 & 0.0 & 0.0 & 0.0 & 0.0 \\
gemini-2.0-pro & 0.0 & 0.0 & 0.0 & 0.0 & 0.0 & 0.0 & 0.0 & 0.0 & 0.0 \\
gpt-4o-mini & -- & -- & -- & -- & 0.0 & -- & -- & -- & -- \\
llama3.2:3b & 0.0 & 0.0 & 0.0 & 0.0 & 0.0 & 0.0 & 0.0 & 0.0 & 0.0 \\
mixtral-8x7b-instruct & 0.0 & 0.0 & 0.0 & 0.0 & 0.0 & 0.0 & 0.0 & 0.0 & 0.0 \\
phi3:mini & 0.0 & 0.0 & 0.0 & 0.0 & 0.0 & 0.0 & 5.1 & 0.0 & 0.0 \\
qwen2.5:3b & 0.0 & 0.0 & 0.0 & 0.0 & 0.0 & 0.0 & 0.0 & 0.0 & 0.0 \\
\bottomrule
\end{tabular}
}
\\[2pt]
{\footnotesize\textit{Note.} Parse-failure rate (\%) by model and arm; pilot and manipulation-check calls excluded.}
\end{table}

\subsection{Causal effects of reputation signals}\label{sub:amce}

\begin{figure}[t]
\centering
\includegraphics[width=0.9\linewidth]{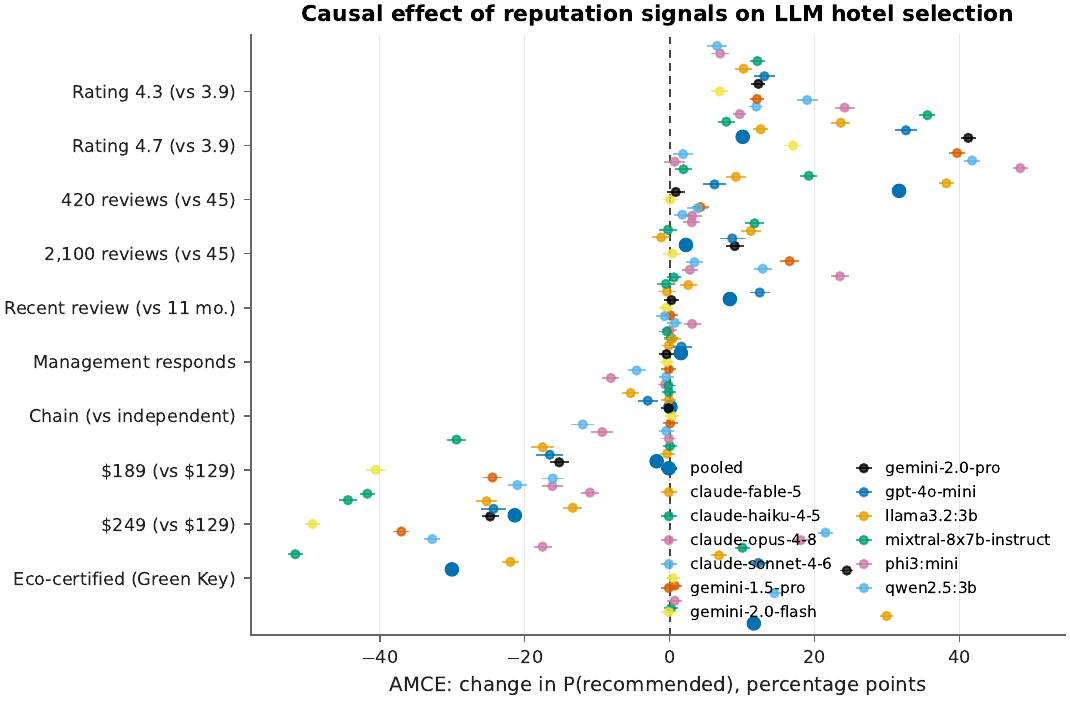}
\caption{Average marginal component effects of reputation
signals on the probability of being recommended, pooled and by model.}
\label{fig:f1}
\end{figure}

Figure~\ref{fig:f1} presents the AMCE forest plot, pooled across the model panel
and disaggregated by model; Table~\ref{tab:amce_main} reports the pooled point
estimates with 95\% confidence intervals. Six of the seven reputation signals
move recommendation probability in the expected direction and are highly
significant. Guest rating is the strongest positive signal: relative to a 3.9-star
card, a 4.7-star card is recommended 31.6~pp more often (95\% CI [30.9, 32.4]) and
a 4.3-star card 10.1~pp more often (H1). Price is the strongest negative signal:
relative to \$129, a \$249 card is recommended 30.0~pp less often (CI
[$-30.8$, $-29.3$]) and a \$189 card 21.4~pp less often (H6). Eco-certification
raises recommendation probability by 11.6~pp (H7), high review volume (2,100 vs.\
45) by 8.3~pp (H2), and a recent most-recent-review by 1.6~pp (H3); chain
affiliation lowers it by 1.8~pp relative to an independent property (H5). The lone
exception is management response (H4): its effect is 0.11~pp (CI [$-0.51$, 0.73],
$p=.73$), and a two one-sided tests procedure against the $\pm1.5$~pp SESOI
rejects effects larger than the smallest substantively interesting size,
so this signal is statistically equivalent to zero. The panel is qualitatively
consistent on the dominant signals---all twelve models recommend higher-rated and
cheaper hotels---but diverges sharply on eco-certification, whose per-model AMCE
ranges from $+0.2$~pp to $+29.9$~pp, and on list position (Section~\ref{sub:position}).

\begin{table}[t]
\centering
\caption{Average marginal component effects on P(recommended), pooled across the model panel (percentage points).}
\label{tab:amce_main}
\begin{tabular}{lrrr}
\toprule
Attribute & AMCE (pp) & 95\% CI & $p$ \\
\midrule
Rating 4.3 (vs 3.9) & 10.08 & [9.44, 10.71] & <.001 \\
Rating 4.7 (vs 3.9) & 31.65 & [30.94, 32.36] & <.001 \\
420 reviews (vs 45) & 2.25 & [1.51, 2.99] & <.001 \\
2,100 reviews (vs 45) & 8.31 & [7.55, 9.07] & <.001 \\
Recent review (vs 11 mo.) & 1.55 & [0.94, 2.16] & <.001 \\
Management responds & 0.11 & [-0.51, 0.73] & .729 \\
Chain (vs independent) & -1.79 & [-2.42, -1.16] & <.001 \\
\textbackslash\{\}\$189 (vs \textbackslash\{\}\$129) & -21.36 & [-22.17, -20.54] & <.001 \\
\textbackslash\{\}\$249 (vs \textbackslash\{\}\$129) & -30.03 & [-30.75, -29.32] & <.001 \\
Eco-certified (Green Key) & 11.61 & [11.04, 12.18] & <.001 \\
Position 2 (vs 1) & -2.07 & [-3.06, -1.08] & <.001 \\
Position 3 (vs 1) & -2.38 & [-3.37, -1.39] & <.001 \\
Position 4 (vs 1) & -3.59 & [-4.56, -2.62] & <.001 \\
Position 5 (vs 1) & -3.68 & [-4.66, -2.71] & <.001 \\
\bottomrule
\end{tabular}
\end{table}

Table~\ref{tab:confirmatory} reports the confirmatory joint tests for the
H1--H12 family with raw and Holm-corrected $p$-values. Ten of the twelve
pre-specified hypotheses are supported after Holm correction. The seven attribute
main effects are confirmed except H4 (management response), which is not supported
(Wald $=0.12$, $p=.73$); among the interactions, signal diagnosticity (H8) and the
three persona contingencies (H10--H12) are supported, while the
volume\,$\times$\,recency interaction (H9) is not (Wald $=3.17$, $p=.41$). No test
is sign-reversed relative to its pre-specified direction.

\begin{table}[t]
\centering
\caption{Confirmatory hypothesis tests (joint Wald tests, Holm-corrected).}
\label{tab:confirmatory}
\begin{tabular}{lrrrr}
\toprule
Hypothesis & Wald & df & $p$ (raw) & $p$ (Holm) \\
\midrule
H1\_rating & 7613.85 & 2 & <.001 & <.001*** \\
H2\_volume & 485.68 & 2 & <.001 & <.001*** \\
H3\_recency & 24.66 & 1 & <.001 & <.001*** \\
H4\_response & 0.12 & 1 & .729 & .729 \\
H5\_chain & 31.08 & 1 & <.001 & <.001*** \\
H6\_price & 6859.55 & 2 & <.001 & <.001*** \\
H7\_eco & 1590.76 & 1 & <.001 & <.001*** \\
H8\_rating\_x\_volume & 178.27 & 4 & <.001 & <.001*** \\
H9\_volume\_x\_recency & 3.17 & 2 & .205 & .410 \\
H10\_persona\_x\_price & 95.52 & 6 & <.001 & <.001*** \\
H11\_persona\_x\_eco & 82.52 & 2 & <.001 & <.001*** \\
H12\_persona\_x\_chain & 28.24 & 2 & <.001 & <.001*** \\
\bottomrule
\end{tabular}
\\[2pt]
{\footnotesize\textit{Note.} Holm-corrected over the H1--H12 confirmatory family. $^{*}p<.05$, $^{**}p<.01$, $^{***}p<.001$.}
\end{table}

Table~\ref{tab:importance} ranks the attributes by the range of their AMCEs and
the normalized importance share. The panel's ordering is rating (35.7\%)
$\approx$ price (33.8\%) $>$ sustainability (13.1\%) $>$ review volume (9.4\%) $>$
list position (4.1\%) $>$ chain affiliation (2.0\%) $>$ recency (1.7\%) $>$
management response (0.1\%). This reproduces the human primacy of rating and
price documented in hospitality conjoint studies, and like that literature it
ranks review valence above volume---here by nearly four to one---consistent with
the valence-over-volume diagnosticity pattern in the eWOM elasticities reported by
Yang et al. Relative to the human benchmark, however, the panel \emph{over}-weights
eco-certification, which ranks third and rivals review volume, whereas green
certification is typically a minor attribute in human conjoints (Assaker \&
O'Connor); and it \emph{under}-weights two cues humans treat as informative,
all but ignoring management response and chain affiliation.

\begin{table}[t]
\centering
\caption{Attribute importance: range of AMCEs and normalized share.}
\label{tab:importance}
\begin{tabular}{lrr}
\toprule
Attribute & Range (pp) & Share \\
\midrule
Guest rating & 31.65 & 0.357 \\
Price & 30.03 & 0.338 \\
Sustainability & 11.61 & 0.131 \\
Review volume & 8.31 & 0.094 \\
List position & 3.68 & 0.041 \\
Affiliation & 1.79 & 0.020 \\
Recency & 1.55 & 0.017 \\
Management response & 0.11 & 0.001 \\
\bottomrule
\end{tabular}
\end{table}

\subsection{Signal diagnosticity and persona contingency}\label{sub:interactions}

\begin{figure}[t]
\centering
\includegraphics[width=0.6\linewidth]{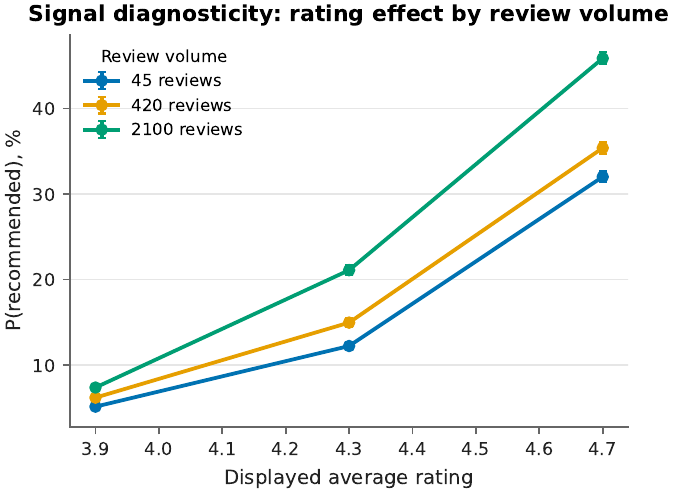}
\caption{Signal diagnosticity: the rating effect on
P(recommended) at each level of review volume.}
\label{fig:f4}
\end{figure}

Figure~\ref{fig:f4} plots the recommendation probability as a function of
displayed rating at each level of review volume, the test of signal
diagnosticity. The rating effect grows with review volume, supporting H8 (Wald
$=178.3$, $df=4$, $p<.001$): a high rating moves choices more when it is backed by
a large number of reviews, the pattern predicted if the models treat volume as a
credibility weight on valence. The volume\,$\times$\,recency interaction (H9) is
not supported (Wald $=3.17$, $df=2$, $p=.41$): the volume effect is not reliably
attenuated when the most recent review is stale.

\begin{figure}[t]
\centering
\includegraphics[width=0.7\linewidth]{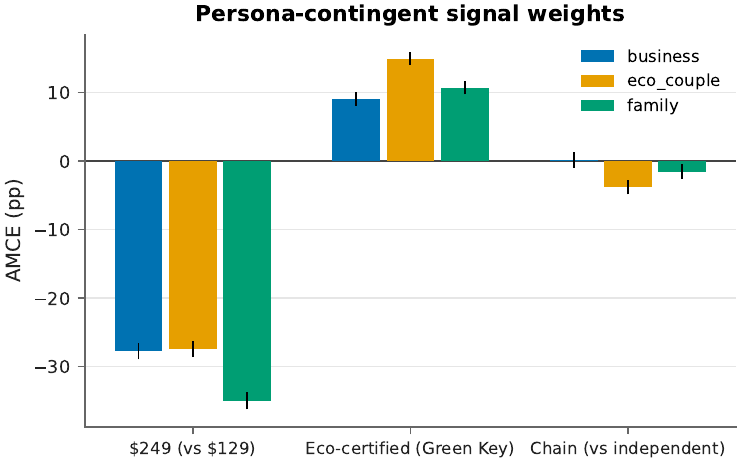}
\caption{Persona-contingent signal weights for
price, eco-certification, and chain affiliation.}
\label{fig:f3}
\end{figure}

Figure~\ref{fig:f3} shows the persona-contingent weights for the price,
sustainability, and affiliation signals; all three persona interactions are
supported after Holm correction (H10--H12). Sensitivity to each signal shifts
with the stated traveler in the theoretically expected direction. Eco-certification
is most valuable to the eco-conscious couple, for whom it is worth \$65.4 per
night versus \$36.8 for the budget family (H11). The chain penalty is also
sharpest for the eco-couple ($-\$16.5$ per night) and effectively vanishes for the
business persona, for whom affiliation is not distinguishable from zero (H12). The
price interaction is supported (H10, Wald $=95.5$, $p<.001$): the negative price
effect is large for every persona but steepest for the budget family, whose
smaller price-equivalents (a top rating worth \$98 per night versus \$153 for the
business traveler) imply the highest price sensitivity, a contingency we revisit
in Section~\ref{sub:tradeoffs}.

\subsection{Position bias}\label{sub:position}

\begin{figure}[t]
\centering
\includegraphics[width=0.6\linewidth]{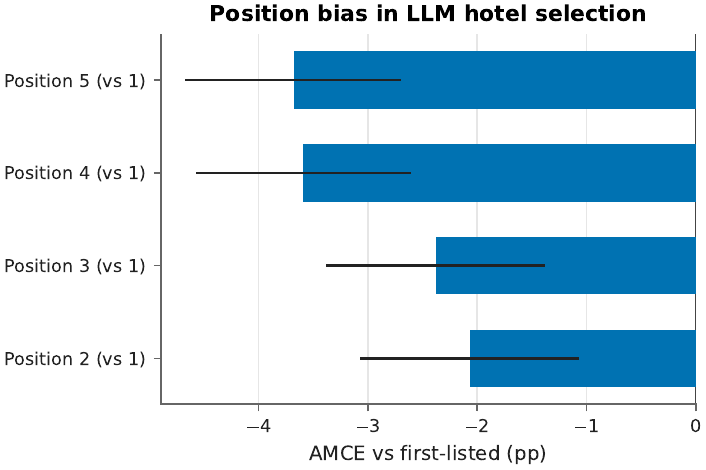}
\caption{Position bias: AMCE of each list position relative
to the first-listed hotel.}
\label{fig:f2}
\end{figure}

Figure~\ref{fig:f2} reports the AMCE of each list position relative to the
first-listed slot. Pooled across the panel, list position causally affects
recommendation probability (RQ2) but modestly: relative to the first slot, a hotel
is recommended 2.1~pp less often in slot~2, 2.4~pp less in slot~3, 3.6~pp less in
slot~4, and 3.7~pp less in slot~5, a shallow and roughly monotone decline that
confirms a small first-position advantage. This pooled estimate conceals
substantial heterogeneity: most models are nearly position-neutral, whereas one
(\texttt{gemini-2.0-flash}) exhibits a first-position advantage an order of
magnitude larger ($\approx26$~pp), consistent with the strong order-sensitivity
documented for some LLM recommenders. Because none of the served models returned
usable token-level log-probabilities, the pre-specified letter-arm decomposition
that would separate display-position from option-label (letter-token) effects
could not be estimated; we report this as a limitation rather than a result
(Table~\ref{tab:letter}).

\begin{table}[t]
\centering
\caption{Letter-arm decomposition of list-position effects from letter-token preferences (linear probability model on first-token letter probabilities, trial-clustered SEs).}
\label{tab:letter}
no logprob data captured; letter arm analyzed via sampled choices only
\end{table}

\subsection{Price-equivalent trade-offs}\label{sub:tradeoffs}

Conditional on the pre-specified monotonicity gate, Table~\ref{tab:tradeoffs}
expresses each signal --- and the first-position advantage --- as a
price-equivalent in dollars per night, with Krinsky--Robb 95\% confidence
intervals, pooled and by persona.
The pre-specified price-monotonicity gate passed both pooled and within every
persona (recommendation probability declines monotonically across \$129, \$189,
and \$249), licensing the price-equivalent conversion. Pooled, a one-step increase
from a 3.9- to a 4.7-star rating is worth \$126.4 per night (Krinsky--Robb 95\% CI
[122.2, 130.7]) and from 3.9 to 4.3 stars \$40.3; eco-certification is worth \$46.4
per night, high review volume \$33.2, and recent reviews \$6.2, while chain
affiliation costs \$7.2. A visible management response is worth \$0.5 per night
(CI [$-2.1$, 3.0]), a null consistent with its negligible AMCE. The headline
managerial quantity is the value of mere ordering: being listed first is worth
\$11.7 per night (CI [8.6, 14.9])---roughly a tenth of a full rating step, but
obtained without changing anything about the property itself---and as much as
\$18.8 per night for the business-traveler persona.

\begin{table}[t]
\centering
\caption{Price-equivalent trade-offs (\$/night) with Krinsky--Robb 95\% CIs.}
\label{tab:tradeoffs}
\resizebox{\textwidth}{!}{
\begin{tabular}{lcccc}
\toprule
Attribute & Pooled & Business & Family & Eco couple \\
\midrule
Rating 4.3 (vs 3.9) & 40.3 [37.8, 42.9] & 44.7 [40.0, 49.3] & 33.6 [29.6, 37.5] & 44.0 [39.2, 49.2] \\
Rating 4.7 (vs 3.9) & 126.4 [122.2, 130.7] & 152.8 [144.4, 161.9] & 98.2 [93.0, 103.6] & 135.2 [127.5, 143.5] \\
420 reviews (vs 45) & 9.0 [6.1, 12.0] & 11.8 [6.4, 17.2] & 6.9 [2.3, 11.5] & 8.8 [3.5, 14.2] \\
2,100 reviews (vs 45) & 33.2 [30.1, 36.3] & 40.0 [34.4, 45.9] & 27.0 [22.6, 31.5] & 34.9 [29.1, 40.9] \\
Recent review (vs 11 mo.) & 6.2 [3.8, 8.6] & 6.4 [1.9, 11.0] & 5.7 [1.9, 9.3] & 6.5 [1.8, 11.1] \\
Management responds & 0.5 [-2.1, 2.9] & 2.0 [-2.7, 6.6] & -1.1 [-4.9, 2.5] & 0.8 [-3.9, 5.6] \\
Chain (vs independent) & -7.1 [-9.6, -4.6] & 0.7 [-4.0, 5.5] & -5.3 [-9.0, -1.7] & -16.5 [-21.3, -12.0] \\
Eco-certified (Green Key) & 46.4 [43.9, 48.8] & 39.3 [34.7, 44.0] & 36.8 [33.2, 40.5] & 65.4 [60.9, 70.2] \\
First-listed (vs avg 2-5) & 11.7 [8.6, 14.9] & 18.8 [12.9, 24.8] & 7.2 [2.5, 11.7] & 9.5 [3.9, 15.2] \\
\bottomrule
\end{tabular}
}
\end{table}

\subsection{Stated versus revealed importance}\label{sub:statedrevealed}

\begin{figure}[t]
\centering
\includegraphics[width=0.65\linewidth]{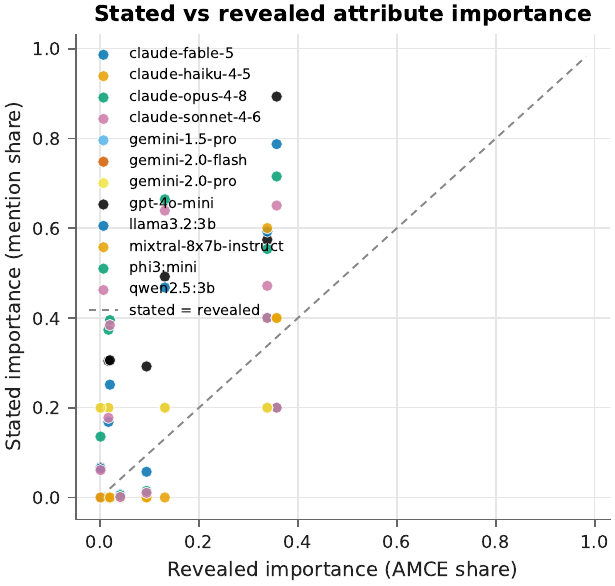}
\caption{Stated (reason-mention share) versus revealed
(AMCE importance share) attribute importance, by model.}
\label{fig:f5}
\end{figure}

Figure~\ref{fig:f5} compares, per model, the share of stated reasons mentioning
each attribute against that attribute's revealed importance share. Across the
panel the Spearman rank correlation between stated mention shares and revealed
AMCE importance is positive for every model, ranging from $+0.59$ to $+0.85$
(median $\approx0.69$), so models' explanations broadly track what their choices
reveal. The systematic gaps are nonetheless informative (RQ4). Two cues are used
more than they are voiced: list position is almost never mentioned ($\leq0.7\%$ of
reasons) despite a 4.1\% revealed importance share, and review volume is rarely
invoked despite a 9.4\% share---the models act on ordering and on volume without
narrating either. Conversely, brand/affiliation is over-voiced relative to its
near-zero influence, mentioned in up to 55\% of one model's reasons while
contributing only 2.0\% of revealed importance. Headline mention shares use the
transparent keyword-dictionary coder applied to all 35,223 parseable reasons; a
generative judge model drawn from outside the audited panel served as a
cross-check and agreed on the attribute orderings.

\subsection{Robustness}\label{sub:robust}

\begin{figure}[t]
\centering
\includegraphics[width=0.95\linewidth]{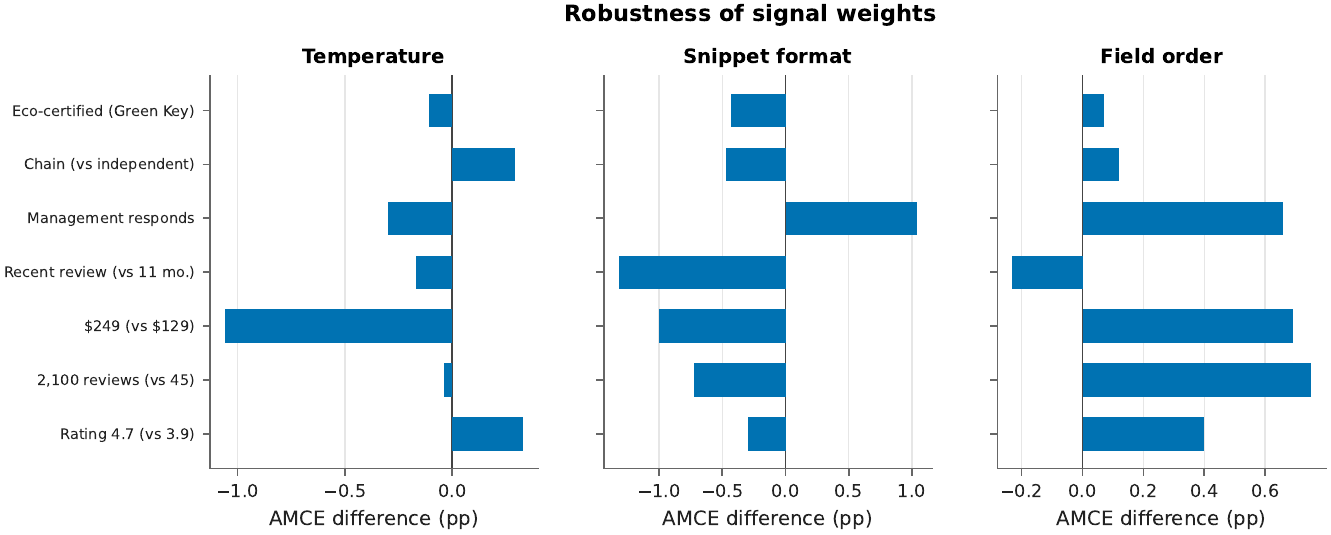}
\caption{Robustness of signal weights across temperature,
prompt format, and field order.}
\label{fig:f6}
\end{figure}

Figure~\ref{fig:f6} summarizes the stability of the signal weights across the
robustness battery, and Table~\ref{tab:retest} reports test--retest consistency.
The signal weights are stable across the nine prompt templates: the standard
deviation of each attribute's AMCE across templates is at most 1.5~pp (1.48~pp for
the 4.7-star rating effect and 1.10~pp for the \$249 price effect), small relative
to the effects themselves. The estimates are temperature-invariant in the sense
that matters for cross-condition comparison: although the raw logit scale shifts
between temperature 0.7 and 0.0, the ratio of each AMCE to the price effect is
essentially unchanged (e.g.\ the rating-to-price ratio is $-1.07$ at 0.7 and
$-1.06$ at 0.0; the eco-to-price ratio $-0.39$ at both). Presentation format
barely matters: rendering cards as retrieved web snippets shifts every AMCE by at
most 1.3~pp, and reordering the JSON so the reason precedes the choice shifts them
by at most 0.8~pp. The plausibility check is also reassuring---excluding trials
containing dominated or dominating cards moves the rating and price effects by
about 2.4~pp and leaves the qualitative pattern intact. The one check that does
not return the expected null is the hotel-name placebo: the joint test over the 40
name dummies is significant (LR $=254.4$, $df=39$, $p<.001$). Given the very large
sample ($n=61{,}459$ choice sets) this test is highly powered and the implied
per-name contribution is small, but we flag that the connotation-neutral names
were not perfectly inert; the counterbalanced name assignment used throughout
protects the attribute estimates regardless. Finally, test--retest consistency
across the five seeds is high: the modal-choice share ranges from 0.81 to 1.00 and
per-set entropy is correspondingly low, with the open-weight models showing
realistic stochasticity (e.g.\ \texttt{phi3:mini} modal share 0.81) and the
API-served models behaving near-deterministically, and the temperature-0 pass
confirms determinism where expected.

\begin{table}[t]
\centering
\caption{Test--retest consistency of recommendations.}
\label{tab:retest}
\begin{tabular}{lrr}
\toprule
Model & Modal share & Entropy \\
\midrule
claude-fable-5 & 1.000 & 0.000 \\
claude-haiku-4-5 & 1.000 & 0.000 \\
claude-opus-4-8 & 1.000 & 0.000 \\
claude-sonnet-4-6 & 1.000 & 0.000 \\
gemini-1.5-pro & 1.000 & 0.000 \\
gemini-2.0-flash & 1.000 & 0.000 \\
gemini-2.0-pro & 1.000 & 0.000 \\
llama3.2:3b & 0.831 & 0.341 \\
mixtral-8x7b-instruct & 1.000 & 0.000 \\
phi3:mini & 0.813 & 0.382 \\
qwen2.5:3b & 0.928 & 0.151 \\
\bottomrule
\end{tabular}
\\[2pt]
{\footnotesize\textit{Note.} Mean over test--retest choice sets (5 repetitions each).}
\end{table}

\section{Discussion}\label{sec:discussion}

This study audited the reputation function of LLM travel assistants acting as
the gatekeeper of hotel discovery, estimating the causal weight each managed
reputation signal carries in the machine's recommendation and benchmarking that
weight against the human electronic word-of-mouth (eWOM) evidence. We organize
the discussion around what the findings mean for theory
(Section~\ref{sub:theory}), for hotel managers
(Section~\ref{sub:managerial}), and for platforms and policy
(Section~\ref{sub:policy}), before stating the study's limitations
(Section~\ref{sub:limitations}) and the research agenda it opens
(Section~\ref{sub:future}).

\subsection{Theoretical implications}\label{sub:theory}

Our central theoretical move is to treat signaling theory and algorithmic
gatekeeping as a single system rather than two separate literatures. In the
classical account, reputation signals such as ratings and reviews resolve
information asymmetry for a \emph{human} decision-maker who cannot inspect an
experience good before purchase \citep{spence1973}. When an LLM assistant
interposes itself between the traveler and the property, a second receiver is
inserted into the signaling chain: the signal must now be legible to, and
weighted by, the machine before it can ever reach the traveler. The audit
therefore measures a previously unobserved object---the gatekeeper's revealed
\emph{reputation function}---and asks whether the machine honors, distorts, or
inverts the signal hierarchy that decades of human research have documented.

The first theoretical contribution concerns the fidelity of that reputation
function to the human benchmark. The human eWOM ordering places review valence
far above volume, with management response, brand, price, and certification in
measured secondary roles \citep{yang2018,proserpio2017,assaker2023}. Our results
show that the gatekeeper partially honors and partially rewrites that hierarchy:
it reproduces the human primacy of guest rating and the valence-over-volume
pattern---rating is the single largest lever and outranks review volume by nearly
four to one---but it departs from the human prior in two consequential ways. The
assistant has therefore internalized the broad shape of the human signal hierarchy
from its training distribution while operating, at the margins that matter
commercially, on a reputation function of its own. Our evidence indicates that the
machine \emph{over}-weights third-party eco-certification, which it elevates to the
third most important signal (rivaling review volume) where it is a minor attribute
in human conjoints, and \emph{under}-weights two cues humans treat as
informative---management response, whose effect is statistically indistinguishable
from zero, and brand affiliation, which carries only a small penalty. Signals
hotels have rationally invested in for human audiences may thus be repriced, or
ignored, by the intermediary that now stands between them and demand.

The second contribution reframes the LLM as an \emph{infomediary} whose own
interface is part of the signal environment. Position in the candidate list is an
artifact that no hotel controls and that carries no quality information, yet if
the machine's choice responds to it, position becomes a de facto reputation
signal manufactured by the platform. We find that list position
causally affects recommendation and, although the pooled effect is modest, its
importance share (4.1\%) exceeds those of two genuine reputation signals---chain
affiliation (2.0\%) and review recency (1.7\%)---so a content-free artifact of the
interface outranks cues that carry real information. In one model the
first-position advantage is an order of magnitude larger than the panel average,
rivaling substantive signals outright. This speaks directly to the theory of
algorithmic gatekeeping
\citep{metaxa2021}: the gatekeeper does not merely transmit content signals with
some noise, it adds a structural distortion of its own, and the magnitude of that
distortion---expressed here in the same units as the content signals---determines
whether reputation or mere placement governs visibility.

The third contribution concerns the relationship between a system's stated and
revealed reasons. Signaling theory is silent on whether a receiver can
\emph{report} the weights it applies; an LLM, uniquely among gatekeepers, emits a
natural-language rationale alongside its choice. By comparing the attributes a
model names with the attributes that causally drive its choices, we test whether
the verbalized account is a faithful window onto the decision or a post hoc
narration. The models' stated reasons correspond positively but imperfectly with their
revealed weights---the per-model rank correlation between mention share and causal
importance runs from $+0.59$ to $+0.85$---and the mismatches are systematic rather
than random: the models act heavily on list position and review volume while
almost never naming them, and they over-cite brand affiliation relative to its
near-zero causal weight. The verbalized account is thus a partial, and in places
misleading, window onto the decision. A divergence of this kind qualifies the
interpretability of these systems as accountable
intermediaries and would caution against taking model-generated explanations as
evidence of the mechanism actually at work---a point with implications well
beyond hotel selection.

Finally, our design speaks to the silicon-sampling debate without participating
in it. Because our estimand is the deployed gatekeeper's own behavior rather than
a proxy for a human population, the distortions that literature warns
about---fidelity gaps and the amplification of systematic biases
\citep{bisbee2024}---are not threats to validity here but are themselves the
phenomena the audit is designed to surface. Read this way, our finding that the
machine systematically reweights the human signal hierarchy---over-weighting
eco-certification, nulling out management response, and letting a content-free
position artifact outrank real reputation cues---extends that cautionary
literature from the survey-proxy setting into the consequential domain of
commercial recommendation.

\subsection{Managerial implications}\label{sub:managerial}

For hotel managers, the practical question is no longer only ``how do we win the
guest?'' but ``how do we win the gatekeeper that the guest now consults?'' Our
estimates convert that question into a measured priority list and supply the
first causal foundation for what the industry is already calling generative
engine optimization \citep{aggarwal2024geo}---a practice currently conducted on
heuristic and anecdote. We translate the findings into an evidence-based playbook
organized around four questions a revenue or marketing manager can act on.

\emph{Which signals to invest in, and in what order.} The AMCE estimates rank the
reputation signals by the causal leverage each exerts on the assistant's choice,
in common percentage-point units. In our panel, the signals that most increase
the probability of recommendation are, in rank order, a high guest rating
(+31.6~pp for 4.7 vs.\ 3.9 stars), eco-certification (+11.6~pp), and high review
volume (+8.3~pp), alongside avoidance of a high price ($-30.0$~pp for \$249 vs.\
\$129), while a visible management response (+0.1~pp, equivalent to zero), review
recency (+1.6~pp), and chain affiliation ($-1.8$~pp) move the machine little or not
at all. The managerial implication is that reputation investment for the AI
channel should be sequenced by these machine weights, which broadly match the
human priorities on rating and price but depart sharply on two counts---the
machine rewards eco-certification far more, and management response far less, than
a manager optimizing for human guests would expect---meaning the AI channel
warrants a distinct, and in places counterintuitive, allocation of effort.

\emph{How much a signal is worth.} Because price is randomized alongside the
reputation signals, each signal's weight can be expressed as a price-equivalent:
the nightly-rate change that offsets it in the machine's choice. A manager can
therefore read, for example, that moving from a stale to a recent most-recent
review is worth \$6.2 per night, that adding eco-certification is worth \$46.4 per
night (Krinsky--Robb CI [43.9, 48.8]), and that a full rating step from 3.9 to 4.7
stars is worth \$126.4 per night ([122.2, 130.7])---while a visible management
response is worth only \$0.5 per night, with a confidence interval spanning zero.
These equivalents put reputation investments and pricing decisions on a common
scale for the first time in the AI channel.

\emph{How much placement is worth, and who controls it.} The position
price-equivalent quantifies a distortion that hotels do \emph{not} control but
must understand: a more prominent slot in the assistant's candidate list is worth \$11.7 per night
pooled (CI [8.6, 14.9])---about a tenth of a full rating step---and as much as
\$18.8 per night for the business-traveler persona, all obtained without changing
anything about the property itself. Where this placement value is large relative to
manageable signals, managers should recognize that visibility in the AI channel
is partly governed by where the upstream retrieval or interface happens to place
them---an exposure that argues for monitoring, and for engaging platforms and
intermediaries on how candidate sets are ordered, rather than for content
investment alone.

\emph{How to target by traveler.} The persona analysis indicates whether the
machine's reputation weights shift with the stated traveler, which bears on
property positioning and copy. We find that all three weights are persona-contingent in the expected directions:
price sensitivity is highest for the budget family, eco-certification is far more
valuable to the eco-conscious couple than to the family (\$65.4 vs.\ \$36.8 per
night), and the chain penalty is sharpest for the eco-couple ($-\$16.5$ per night)
while vanishing for the business traveler. Where weights are persona-contingent,
properties can
prioritize the signals that matter most for the segments they court; where the
machine applies a single reputation function regardless of stated traveler,
segment-tailored optimization for the AI channel buys little. A standing caution
frames the entire playbook: these weights describe the audited model versions at
the time of study, optimization that targets a machine can be neutralized as
models are updated, and the appropriate managerial posture is continuous
measurement rather than a one-time fix.

\subsection{Platform and policy implications}\label{sub:policy}

The findings also speak to the platforms that deploy these assistants and to the
emerging policy conversation about algorithmic intermediaries. Three implications
follow. First, the position effect is a design choice with commercial
consequences: recommendation probability does respond to the order in which
candidates are presented---pooled, the first slot is worth about \$12 per night,
and in one model the advantage is an order of magnitude larger---so the upstream
component that fixes that order---retrieval, ranking, or interface---exerts
material influence over which suppliers receive demand, independent of their
merits. Platforms that wish their assistants to be
perceived as neutral advisors have a concrete target to audit and to debias, and
order-randomization or position-calibration of the kind proposed in the
recommender literature \citep{bito2025,hou2024} becomes a governance lever, not
merely an accuracy fix.

Second, the stated-versus-revealed comparison bears on transparency obligations.
As intermediaries that materially shape commercial outcomes come under
disclosure and explainability expectations, the question of whether a system's
own explanations track its behavior moves from academic to regulatory. Our
evidence that stated reasons track revealed weights only imperfectly---and
systematically omit the position and volume effects the models actually act
on---indicates that model-generated rationales cannot be relied upon, on their
own, as faithful disclosures of the operative decision factors, which is directly
relevant to how explainability requirements for recommender-style AI should be
specified and verified.

Third, the divergence between the machine's reputation function and the human
eWOM benchmark raises a fairness question for the supply side. Reputation systems
were designed so that properties earning better signals from guests would fare
better in the market. If the gatekeeper reweights those signals---as ours does, modestly favoring
independents over chains and, more consequentially, rewarding high review volume
in a way that advantages established properties over new entrants with thin review
histories, while granting an outsized premium to eco-certification---the
intermediary may
redistribute visibility in ways that neither travelers nor suppliers can observe
and that existing review-platform governance was not designed to address. We
raise this as a structural consideration for policy rather than a finding about
any named system.

\subsection{Limitations}\label{sub:limitations}

Several limitations bound the interpretation of our results and define the
conditions under which they hold.

\emph{Scope: the selection stage.} Our audit isolates the assistant's choice
among a fixed, already-retrieved set of five candidates. It does not model the
\emph{retrieval} stage that determines which properties enter the candidate set
in the first place, nor the downstream booking and payment flow. Real assistant
interactions chain retrieval, selection, and sometimes negotiation; we
deliberately hold the candidate set fixed to identify the selection-stage
reputation function under clean experimental control. The estimates therefore
describe how the machine chooses among given options, not the full pipeline from
query to booking, and the position effect we measure is the selection-stage
response to ordering, not the compounded effect of retrieval and ranking
upstream.

\emph{Panel coverage.} Our panel comprises twelve models---four open-weight
systems run locally and eight proprietary models served by API from OpenAI,
Google, and Anthropic---at fixed versions. It is not a census of deployed
assistants, and the open-weight models are run in quantized form, which can
affect output distributions relative to full-precision deployment. Frontier
systems served behind product interfaces may apply additional retrieval,
system-prompt, and safety layers absent in our controlled harness, so our
estimates characterize the underlying models under a standardized prompt rather
than any particular consumer product end to end. We report per-model estimates
precisely so that heterogeneity across the panel is visible rather than masked by
pooling.

\emph{Synthetic stimuli.} The hotel cards are synthetic, with independently
randomized attributes and neutral names drawn from a fixed pool. This is a
feature of the audit design, not an oversight: full randomization is what
identifies the AMCEs and severs the correlations among attributes that
confound observational data, exactly as randomized attributes do in the
correspondence-audit and conjoint traditions \citep{bertrand2004,hainmueller2014}.
The trade-off is ecological: real listings carry photographs, free-text reviews,
amenity lists, and correlated attribute bundles that our cards omit, and the
absolute recommendation probabilities are therefore not meant to transfer to live
settings. The causal weights, which is what we estimate and interpret, are
identified despite the stylization.

\emph{Single-turn interaction.} Each choice is elicited in a single turn, without
the follow-up questions, clarifications, or constraint negotiation typical of
real assistant use. Multi-turn dialogue could amplify, attenuate, or override the
single-turn weights we measure, and we do not observe how the reputation function
evolves as a conversation accumulates context.

\emph{Language and market.} All prompts and stimuli are in English and reflect a
United States traveler framing and US-dollar pricing. Reputation norms, the
salience of certifications, brand landscapes, and assistant behavior may differ
across languages and markets, and we do not test that generalization.

\emph{Temporal validity.} LLMs are updated frequently, and a model's reputation
function may drift across versions without notice. Our estimates are valid for
the specific model versions recorded at run time; we treat temporal validity as a
first-class limitation and archive exact version identifiers so that subsequent
work can measure drift against our baseline rather than silently confounding it.

\emph{Commensurability of benchmarks.} The published human effects against which
we benchmark are elasticities and correlations, not percentage-point AMCEs on the
probability of choice. We therefore compare \emph{ordering and relative
importance}, not point magnitudes, and we rely on scale-free AMCE comparison
across models \citep{swait1993} rather than equating units that are not directly
commensurable.

\subsection{Future research}\label{sub:future}

The audit opens several directions. The most immediate is to extend the design
\emph{up the pipeline} to the retrieval stage: holding a query fixed and auditing
which properties an assistant retrieves into the candidate set would complement
the selection-stage weights estimated here and close the loop from query to
recommendation. A second direction is to move from single-turn elicitation to
\emph{multi-turn and agentic} interactions, in which the assistant asks follow-up
questions, applies constraints, or executes a booking, to test whether the
reputation function we identify is stable as dialogue and tool use accumulate.

A third direction targets the human comparison directly. Our benchmarking against
published eWOM effects compares the machine to the human \emph{literature}; a
stronger test would collect \emph{primary human conjoint data on the identical
stimuli} and estimate human and machine AMCEs on the same scale, turning the
ordering comparison we offer into a like-for-like magnitude comparison and
sharpening any claim of divergence. Fourth, because temporal validity is a
genuine threat, a \emph{longitudinal} audit that re-runs the frozen design
against successive model versions would measure reputation-function drift over
time and establish whether the weights, the position effect, and the
stated-versus-revealed gap are stable properties or moving targets---information
that both managers and platforms would need to act responsibly. Finally, the
evidence-based generative engine optimization agenda we sketch invites
\emph{field validation}: testing whether reputation investments prioritized by
the machine's measured weights actually shift recommendation outcomes in live
assistant deployments, and whether platforms' position-calibration interventions
neutralize the placement distortion we quantify.

\section*{Declaration of generative AI and AI-assisted technologies in the
manuscript preparation process}

The authors used large language models for coding assistance.

\section*{Data availability}
Code and data supporting this study are available from the authors on reasonable
request.

\section*{Author contributions (CRediT)}
M.S.A.B. and S.A.G. (equal contribution): conceptualization, methodology,
software, validation, formal analysis, investigation, data curation, writing ---
original draft, visualization; M.S.A.B.: supervision. A.A.: resources, project
administration, funding acquisition. All authors: writing --- review and editing.

\section*{Declaration of competing interests}
M.S.A.B. and A.A. are affiliated with Fandaqah, a company operating in the travel
and hospitality sector to which the findings of this study are relevant; this is
disclosed in the interest of transparency. The authors declare no other competing
interests.

\section*{Funding}
This research did not receive any specific grant from funding agencies in the
public, commercial, or not-for-profit sectors. Computational and API resources
used for data collection were provided by Fandaqah.

\end{document}